\documentclass{article}

\usepackage{graphicx, color, clrscode}

\usepackage{geometry}
\title{An Evolutionary Approach towards Clustering Airborne Laser Scanning Data}
\author{Ronald Hochreiter \and Christoph Waldhauser}
\date{June 2012}

\begin{document}

\maketitle

\begin{abstract}
In land surveying, the generation of maps was greatly simplified with the introduction of orthophotos and at a later stage with airborne LiDAR laser scanning systems. While the original purpose of LiDAR systems was to determine the altitude of ground elevations, newer full wave systems provide additional information that can be used on classifying the type of ground cover and the generation of maps. The LiDAR resulting point clouds are huge, multidimensional data sets that need to be grouped in classes of ground cover. We propose a genetic algorithm that aids in classifying these data sets and thus make them usable for map generation. A key feature are tailor-made genetic operators and fitness functions for the subject. The algorithm is compared to a traditional k-means clustering. 
\end{abstract}


\section{Introduction}

In land surveying, the identification of structural features covering the surveyed lands are a key issue. Due to the earth's vastness, traditional surveying methods have long been replaced by remote sensing applications. In it's latest iteration, remote sensing uses LiDAR (Light Detection And Ranging) laser beams to detect and record geographical features. While used traditionally in land-based scenarios, ie mounted on the back of a pick-up truck, the LiDAR principle has become airborne. By combining airborne platforms like planes or helicopters with traditional LiDAR equipment, large and otherwise inaccessible areas can be surveyed in comparably little time.

Data as produced by LiDAR scanners is basically a point cloud. With little additional work, these point clouds are geo-referenced in a post-processing step, that makes them displayable in GIS systems. In addition to that, an entire host of post-processing filters and transformations are applied to correct for anomalies in the flight path and stray reflections generated by oblique surfaces. Besides the point coordinates, other features of these points are recorded as well. Using the latest generation of scanners, full wave data can be obtained and analyzed. Key features here include the amplitude and the with of the returned laser signal \cite{wehr1999airborne}.

The resulting point cloud is represented by a rather low-dimensional feature vector, including each point's three dimensional coordinates and the characteristics of the signal returned by each point. While ALS-generated data originally was intended to replace the work-intensive method of stereo-photogrammetry to assess the altitude of geographical features, many more applications are conceivable.

One task of land surveying that is extremely time consuming is the classification of strips of land according to ortho-photographs. This task entails the visual inspection of large amounts of photographs by trained professionals and then the deriving of maps from these photos. Given that the reflective characteristics of laser beams off any surface depends on the material and structure of that surface, LiDAR generated full wave data can be used to classify the point cloud into multiple classes and thus aid in the production of maps.

The specific challenges when classifying ALS point clouds result on one hand from their sheer size \footnote{Full Wave Data is encoded as roughly 10 bytes per point. Given a moderate resolution of 5 points per square meter, results in about 5 Terabytes of data, for a little speck of land like Austria.} and the requirements of obtaining homogeneous areas of reasonable sizes for map presentations.

The size of the point clouds prohibit traditional clustering methods on an entire data set. However, using the principles of machine learning, select subsets, either randomly or by human, can be used in stead of the entire point cloud. This, of course, always bears the risk of not learning from a representative sample of the entire point cloud. As an alternative, it is conceivable that laser return characteristics are the same anywhere on the planet, i.e. a tree always reflects light in a certain way and that these reflections always differ from those of stone, no matter where the measurement took place. In that case, a large amount of points could be analyzed once, and then applied to all point clouds ever recorded.

A larger problem is the smoothing of the detected classification. For mapping applications, it is of little interest if there are actual small patches of grass between the trees of a forest. The entire region should be classified as a forest. As a result of that requirement, it is necessary to adopt some sort of majority vote among neighboring points so as to even out the returned classification. 

A traditional approach would use k-means clustering and a k-nearest neighbor search to arrive at a classification and smoothing, respectively. This traditional approach has a number of shortcomings. For one, k-means clustering settles quickly for local optima, thus not finding necessarily good solutions to classification problem. Further, by using euclidean distance as a metric, k-means searches for equally sized spheroids in all clustered dimensions. While this property works fine with certain data sets, ALS point clouds, unfortunately, are not among them. Here, k-means has serious difficulties differentiating between trees and buildings, for instance. While similar in one dimension, i.e. height, k-means ``obsession'' with equally sized clusters turns rare buildings into more common trees.

Another serious drawback of the traditional approach is the sequential performing of clustering and smoothing. As the classification of regions for maps requires homogeneous regions, it is crucial for most neighboring points to actually belong to the same class, regardless of the signal characteristics of some outlaying points. At the same time, however, the classification should be ignorant of the planar coordinates of any point, as to produce a classification that is mappable to other points in other data sets. As a consequence, smoothing is separated into a distinct step. The employed k nearest neighbor search is painfully slow, as all points need to be traversed. Also, the number of neighbors to specify is often not clear.

This paper suggests a genetic algorithm that finds a partitioning of the point cloud such that ground cover features can be easily identified. The proposed algorithm is by no means the first to attempt clustering by using a genetic approach. However, in deviating from the main stream of research like \cite{maulik2000genetic}, our algorithm considers each point's class membership as the genome.

This paper is organized as follows. Section \ref{sec:algorithm} describes the algorithm in detail, while Section \ref{sec:results} provides promising numerical results. Section \ref{sec:conclusion} concludes the paper.

\section{Algorithm}
\label{sec:algorithm}

The aim of the introduced algorithm is to provide a classification of measured points or rather echoes into types of ground cover. The algorithm seeks to find a solution that generates homogeneous planar areas that are suitable for maps. To this end a genetic algorithm was developed that optimizes a fitness function that takes the quality of the classification and the planar structure of that classification into account. In the following, we will present the details of the used algorithm and discuss the fitness function.

As in any genetic algorithm, genomes are recombined to produce offspring. Each offspring is evaluated with respect to its fitness. The fittest genomes reproduce again. This sequence is repeated over many generations, while each generation has genomes that perform better than any genome before. In the context of clustering ALS point clouds, a genome is the class membership of every point. Thus, for clustering $n$ points into $k$ classes, each genome will be of length $n$ genes. Table \ref{tab:ea} contains the algorithm's pseudocode.

\begin{table}[h]
  \begin{center}
    \caption{Meta-heuristic: Genetic Algorithm}
    \begin{codebox}
      \li $P \gets GenerateInitialPopulation$
      \li $Evaluate(P)$
      \li \kw{while} termination conditions not met \kw{do}
      \li \> $P'_{rand} \gets Random(| P | \times \rho)$ 
      \li \> $P_{elite} \gets Select(P, Quantile(P, \epsilon))$ 
      \li \> $P'_{xover} \gets Recombine(P_{elite}, P)$
      \li \> $P' \gets P'_{rand} \cup P'_{xover}$
      \li \> $P \gets Mutate(P')$
      \li \> $Evaluate(P)$
      \li \kw{end while}
    \end{codebox}
    \label{tab:ea}
  \end{center}
\end{table}

The complexities of real evolution are modeled in our algorithm by using the principles of elite keeping, elite reproduction and a form of fitness proportionate selection. 
  
\subsection{Data}
\label{sec:data}

To evaluate the performance of the algorithm a small ALS point cloud was clustered using the proposed genetic algorithm and with k-means for reference. The point cloud subjected to this treatment represents a small part of Sch\"{o}nbrunn palace gardens in Vienna. Features contained within consist of water surfaces, low and high vegetation, some buildings and gravel paths. The first lines of data from this set are shown in Table \ref{tab:data}. It contains 3-D coordinates $(x,y,z)$, a timestamp $t$, as well as a special set of pre-processed data from the laser scanning process.

\begin{table}
\caption{A selection of the ALS data set used for clustering.}
\label{tab:data}
\begin{center}
\begin{tabular}{rrrrrrrrr}
  \hline
 & x & y & z & t & Amp & EW & EID & nE \\ 
  \hline
1 & -1855.57 & 337175.66 & 71.26 & 299158.24 & 20.00 & 4.10 &   2 &   2 \\ 
  2 & -1855.06 & 337175.71 & 71.22 & 299158.24 & 26.00 & 3.80 &   1 &   1 \\ 
  3 & -1854.53 & 337175.76 & 71.23 & 299158.24 & 32.00 & 3.90 &   1 &   1 \\ 
  4 & -1853.97 & 337175.81 & 71.30 & 299158.24 & 42.00 & 4.00 &   2 &   2 \\ 
  5 & -1853.43 & 337175.87 & 71.31 & 299158.24 & 60.00 & 4.00 &   1 &   1 \\ 
  6 & -1852.91 & 337175.91 & 71.29 & 299158.24 & 65.00 & 4.00 &   1 &   1 \\ 
   \hline
\end{tabular}
\end{center}
\end{table}

\subsection{Fitness function}
The fitness function of any genetic algorithm is crucial in adapting that algorithm to a given problem domain. In our algorithm we use a two-folded fitness function $f$ that covers both: the aspects of a good classification in a classical sense and the homogeneous planar areas criterion as mentioned above. The classification evaluation criterion used is the Dunn-Index \cite{pakhira04,handl05,dunn74} of the classification ($D$). In addition, a penalty is applied to solutions that produce very inhomogeneous areas ($P$) The function thus is the linear combination of two elements:
\begin{equation}
  f = D - P
\end{equation}
The Dunn index was designed to capture two aspects of any clustering classification: a large distance between cluster centers and a small distance of points within a cluster. The large the Dunn index becomes for a classification, the better it is. We thus need to maximize the Dunn index. To incorporate the notion of homogeneous areas, a penalty is subtracted for inhomogeneous solutions. This penalty is computed as the number of points that have an inhomogeneous neighborhood. There will always be some points that are required to have inhomogeneous neighborhoods, those located on the borders of otherwise homogeneous areas. But their number should be considerably lower in good solutions. 

\begin{equation}
  P = \sum_{i,k} | \{p | C_{p_i} \neq C_{p_k} \}|
\end{equation}

\subsection{Encoding}

A similar clustering algorithm has been presented in \cite{Hochreiter10}, see also \cite{Hochreiter11}. If we want to separate e.g. $m=10$ values optimally into e.g. $n=2$ clusters, we take a random chromosome with uniform random numbers between $0$ and $1$, which might look as follows:
$$(0.4387,0.3816,0.7655,0.7952,0.1869,0.4898,0.4456,0.6463,0.7094,0.7547)$$
If we map this vector to represent $n$ centers we obtain: $(1,1,2,2,1,1,1,2,2,2)$, i.e. the cluster splitting is set to each $\frac{1}{n}$, in this case all values smaller than $0.5$ represent cluster $1$ and all values larger than $0.5$ are assigned to cluster $2$. In this case, we can use real-valued operators.

\subsection{Evolutionary Operators}
\label{sec:evoop}

The following evolutionary operators are used in our algorithm:

\paragraph{Elitist selection.} A certain number $o_1$ of the best chromosomes of the parent population is automatically added to the next population (without any changes).

\paragraph{Intermediate crossover.} An intermediate crossover with $\alpha=0.5$ is applied to generate $o_2$ new chromosomes. Thereby an elitist selection of the parents is conducted. One parent is chosen from the $\beta_1$ best and the other from the $\beta_2$ best chromosomes.

\paragraph{Flip Mutation.} A number of $o_3$ chromosomes are chosen (with $\beta_3$ elitist selection) and flip-mutated, i.e. $\gamma_1$ genes with an initial value of $c$ are mutated to $1-c$.

\paragraph{Bisection Mutation.} A number of $o_4$ chromosomes are chosen (with $\beta_3$ elitist selection) and bisected, i.e. $\gamma_2$ genes with an initial value of $c$ are mutated to $\frac{c}{2}$.

\paragraph{Random addition.} $o_5$ chromosomes will be generated randomly and added to the new population.

\section{Results}
\label{sec:results}


The aim was to classify the data described in Section \ref{sec:data} into ten $(n=10)$ different groups. The choosing of the amount of groups was informed by the number of different features as listed above and experience. For instance, trees consist of leaves and branches and trunks, each with very specific laser signal return characteristics.

The algorithm was implemented in R \cite{cran}. In addition, the R packages \cite{clValid,proxy} have been used to speed up the computation of distances and to obtain a Dunn Index scoring for each solution. 

\subsection{Evolutionary convergence}


We conducted $8$ experiments to test different evolutionary operator settings. The experimental settings are listed in Table \ref{tab:exp}. We used different types of parameter settings. Experiment 1 uses a balanced mixture of operators, while experiment 2 resembles a plain random search. Experiments 3-5 are only using one out of the three main evolutionary parameters ($o_2$,  $o_3$,  $o_4$) and Experiments 6-8 are overweighing one of the evolutionary parameters.

\begin{table}
\caption{Evolutionary operator settings for the experiments.}
\label{tab:exp}
\begin{center}
\begin{tabular}{cccccc} \hline
Experiment & Elite  $o_1$ & Crossover  $o_2$ & Mutation Flip  $o_3$ & Mutation Bisect  $o_4$ & Random  $o_5$ \\ \hline
1 & 200 & 200 & 200 & 200 & 200 \\
2 & 500 & 0 & 0 & 0 & 500 \\
3 & 200 & 600 & 0 & 0 & 200 \\
4 & 200 & 0 & 600 & 0 & 200 \\
5 & 200 & 0 & 0 & 600 & 200 \\
6 & 200 & 400 & 100 & 100 & 200 \\
7 & 200 & 100 & 400 & 100 & 200 \\
8 & 200 & 100 & 100 & 400 & 200 \\ \hline
\end{tabular}
\end{center}
\end{table}

The initial population size has been set to $n=1000$. The other parameters described in Section \ref{sec:evoop} have been kept fixed to the following values: $\beta_1 = 0.3, \beta_2 = 0.7, \beta_3=0.5, \gamma_1 = 0.1, \gamma_2 = 0.1$.

\begin{figure}
\begin{center}
\begin{tabular}{cc}
\scalebox{0.4}{
\includegraphics{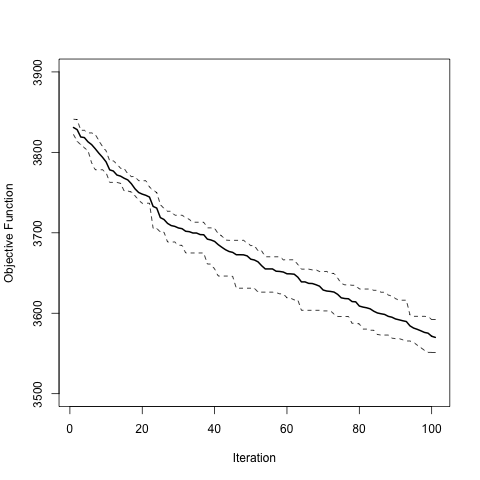}
}
&
\scalebox{0.4}{
\includegraphics{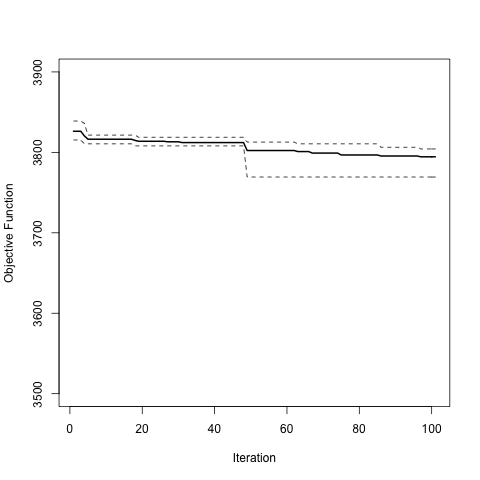}
}
\end{tabular}
\end{center}
\caption{Operator settings: Balanced mixture (left) and plain random search (right).}
\label{fig:res1}
\end{figure}

\begin{figure}
\begin{center}
\begin{tabular}{ccc}
\scalebox{0.2}{
\includegraphics{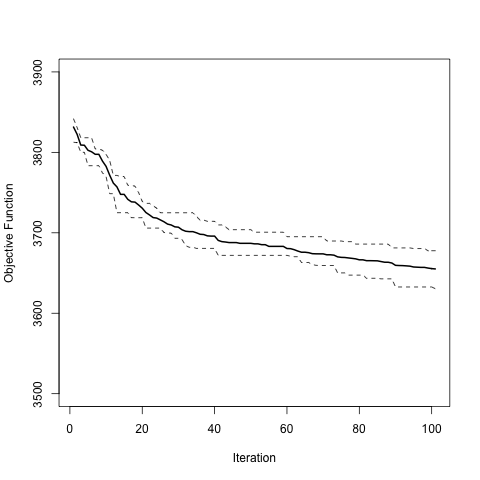}
}
&
\scalebox{0.2}{
\includegraphics{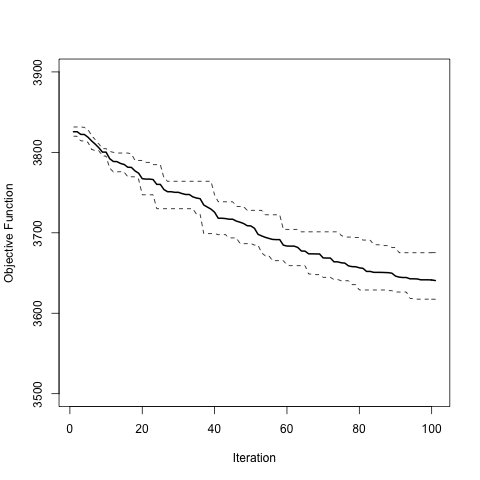}
}
&
\scalebox{0.2}{
\includegraphics{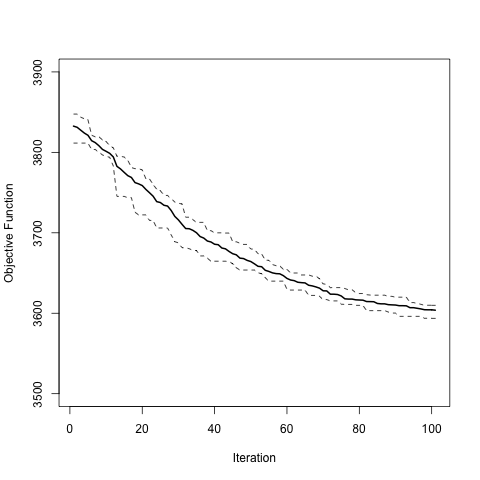}
}
\end{tabular}
\end{center}
\caption{Operator settings: Using only one (out of three) evolutionary operators.}
\label{fig:res2}
\end{figure}

\begin{figure}
\begin{center}
\begin{tabular}{ccc}
\scalebox{0.2}{
\includegraphics{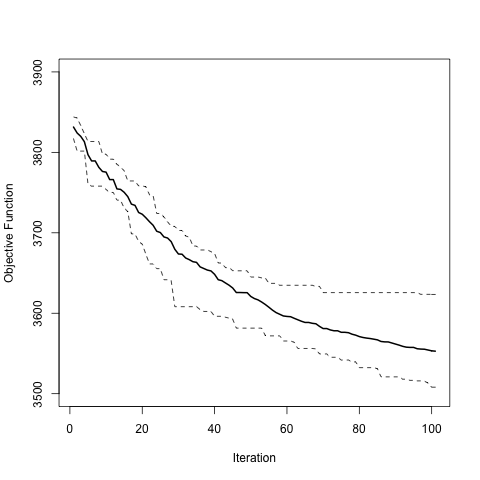}
}
&
\scalebox{0.2}{
\includegraphics{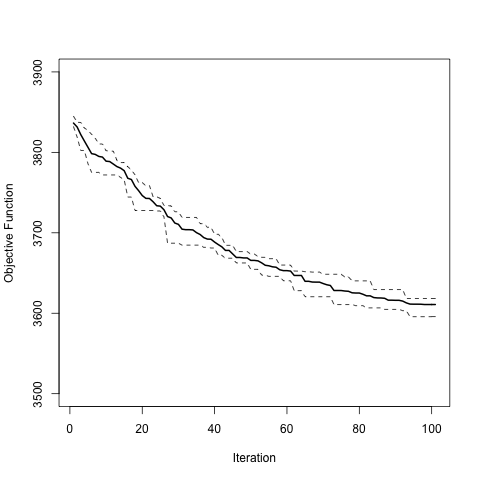}
}
&
\scalebox{0.2}{
\includegraphics{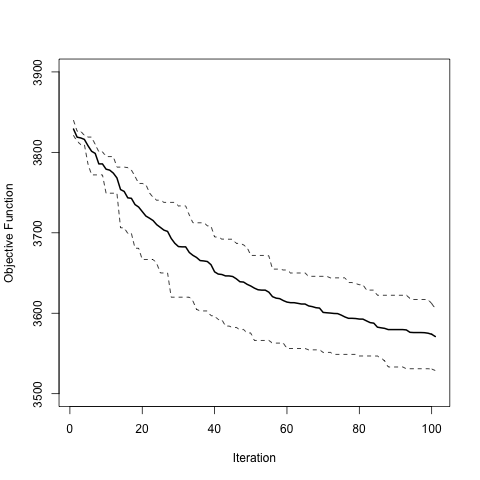}
}
\end{tabular}
\end{center}
\caption{Operator settings: Overweighting one (out of three) evolutionary operators.}
\label{fig:res3}
\end{figure}

We checked the performance of the first $100$ iterations and repeated every experiment $5$ times. Each run (100 iterations) takes approximately five minutes. The results are shown in Fig. \ref{fig:res1}, Fig. \ref{fig:res2}, and Fig. \ref{fig:res3} respectively. A y-axes are scaled to the same dimension. The bold line represents the mean fitness value of the repeated runs and the dashed lines the minimum (and respectively the maximum) of the fitness value of each iteration.

The results show that the evolutionary approach works, as there is no convergence using no evolutionary parameters (Fig. \ref{fig:res1}, right graph). Furthermore, the bisection mutation works very well, but the most stable and least volatile convergence is achieved using a balanced set of operators.

%


\section{Conclusion}
\label{sec:conclusion}


In this paper, we have outlined a new approach to cluster high-dimensional point clouds stemming from airborne laser scanning. Compared to standard k-means clustering the evolutionary approach can be extended to include complex distance measures as well as hybrid evaluators to tackle the complexity of the problem. The results look promising such that future research will focus on improving the algorithm to be scaleable to larger data sets. Another addition is to include local search improvements throughout (and after) the evolutionary optimization process.

\bibliographystyle{plain}
\bibliography{literature}

\end{document}